\documentclass[letterpaper, 10 pt, conference]{ieeeconf}  

\IEEEoverridecommandlockouts                              

\usepackage{graphics} 
\usepackage{epsfig} 
\usepackage{mathptmx} 
\usepackage{times} 
\usepackage{amsmath} 
\usepackage{amssymb}  
\usepackage{xcolor}
\usepackage{booktabs}
\usepackage{bm}

\definecolor{_blue}{RGB}{0,118,186}
\definecolor{_green}{RGB}{31,177,0}
\definecolor{_pink}{RGB}{255,66,161}

\title{\LARGE \bf
Tactile-Based Human Intent Recognition for Robot Assistive Navigation
}

\author{Shaoting Peng$^{1*}$, Dakarai Crowder$^2$, Wenzhen Yuan$^2$, Katherine Driggs-Campbell$^1$
\thanks{$^*$ Corresponding author: {\tt\small peng33@illinois.edu}}
\thanks{$^{1}$Electrical \& Computer Engineering, University of Illinois Urbana-Champaign, IL, 61801, USA}
\thanks{$^{2}$Siebel School of Computing and Data Science, University of Illinois Urbana-Champaign, IL, 61801, USA}
}

\begin{document}

\maketitle

\thispagestyle{empty}
\pagestyle{empty}

\begin{abstract}
Robot assistive navigation (RAN) is critical for enhancing the mobility and independence of the growing population of mobility-impaired individuals. However, existing systems often rely on interfaces that fail to replicate the intuitive and efficient physical communication observed between a person and a human caregiver, limiting their effectiveness. In this paper, we introduce Tac-Nav, a RAN system that leverages a cylindrical tactile skin mounted on a Stretch 3 mobile manipulator to provide a more natural and efficient interface for human navigational intent recognition. To robustly classify the tactile data, we developed the Cylindrical Kernel Support Vector Machine (CK-SVM), an algorithm that explicitly models the sensor's cylindrical geometry and is consequently robust to the natural rotational shifts present in a user's grasp. Comprehensive experiments were conducted to demonstrate the effectiveness of our classification algorithm and the overall system. Results show that CK-SVM achieved superior classification accuracy on both simulated (97.1\%) and real-world (90.8\%) datasets compared to four baseline models. Furthermore, a pilot study confirmed that users more preferred the Tac-Nav tactile interface over conventional joystick and voice-based controls.

\end{abstract}

\section{INTRODUCTION}

Robot assistive navigation (RAN) -- the task of a robot providing physical support while people moving from one place to another -- is of critical importance for people with mobility impairments~\cite{351167}. In the U.S., 12.2\% of adults live with a mobility disability, and the aging population suggests an increasing need for navigation assistance \cite{CDC2023DisabilityImpacts, Freedman2021NHATS}. To address mobility challenges and increase the independence of these populations, researchers have been focusing on developing assistive robotic systems that can help navigation in both indoor and outdoor environments, such as augmented canes \cite{Pyun2013AdvancedAW, Slade2021MultimodalSI}, wheelchairs \cite{351167, PERRIN20101246} and walkers~\cite{Morris2003ARW, 10.3389/fnbot.2020.575889}.  Compared to a cane or wheelchair, walker devices provide adequate physical support while maintaining a degree of user independence, and are also helpful for building endurance and regaining muscle strength, especially during recovery from surgery, injury, or illness \cite{Sehgal2021Mobility, Resnik2009Perspectives}. 
In this work, we explore interaction mechanisms for a robotic walker, which often takes the form of a mobile robot with a handle~\cite{DRAGON}, for assisted navigation.

In the development of effective RAN, a key challenge is human intent recognition, which is fundamental to ensuring the safety and fluidity of the human-robot interaction. The ability to accurately and efficiently detect or input a user's movement intent is highly reliant on the sensing and user interface. Existing interfaces range from simple button arrays and joysticks~\cite{808948} to more complex biosignal sensors like electromyography (EMG) or electroencephalography (EEG) \cite{PERRIN20101246}. However, each approach presents trade-offs. A button array, while intuitive, may divert the user's attention from their surroundings, whereas an EEG sensor, while avoiding physical distraction, can be prone to signal noise and require extensive calibration. Our goal in this work is to mitigate the intent recognition gap by designing an algorithm and system that is natural, accurate, and robust, fostering a human-centric interaction where the user feels they are collaborating with a partner rather than merely operating a machine.

\begin{figure}[!t]
    \centering
    \includegraphics[width=\columnwidth]{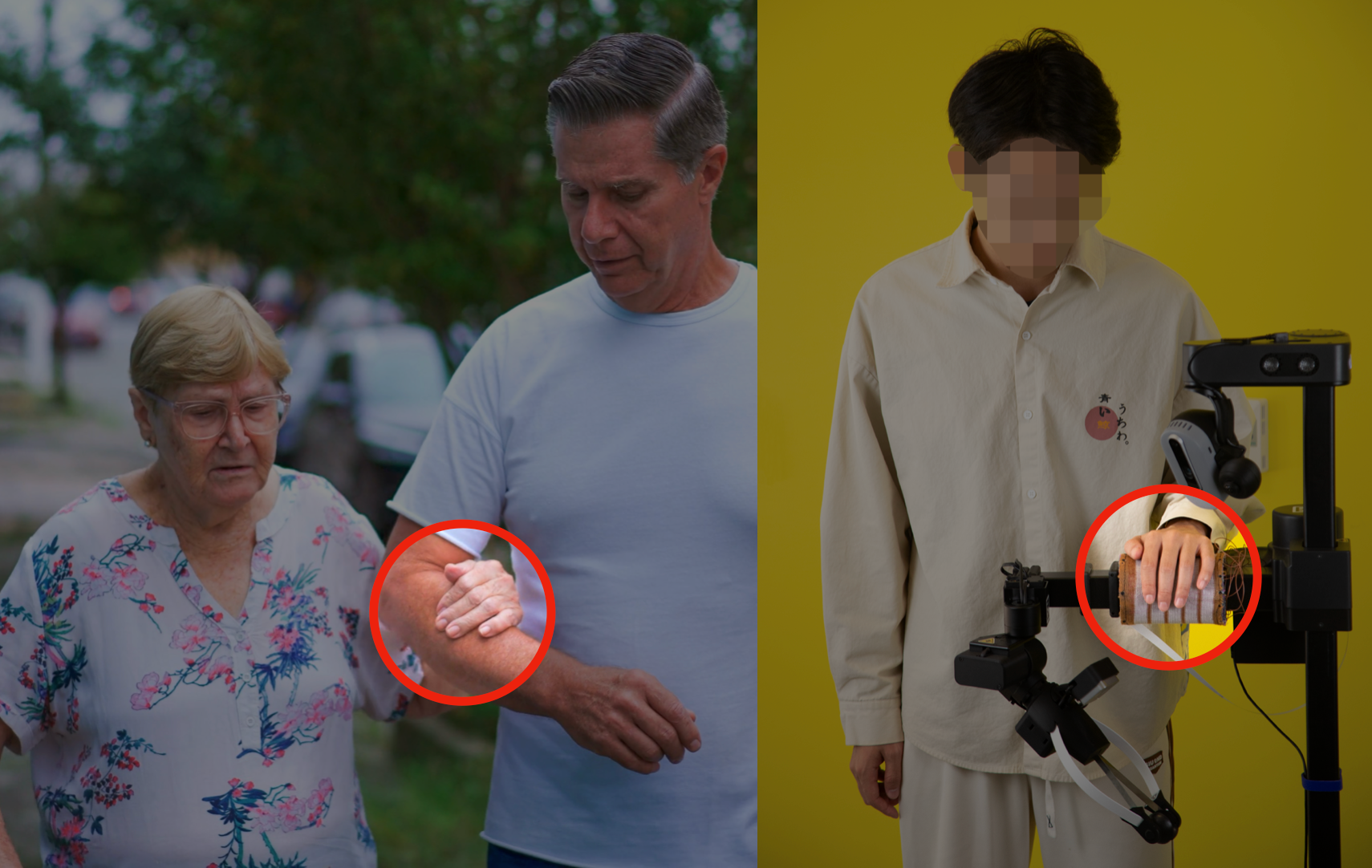}
    \caption{\textbf{Intuition of our work:} Mobility-impaired people convey their moving intents by applying specific grasping force patterns on the arm of caregiver. We adapt this intuitive and natural way of interaction to the robot assistive navigation by putting a tactile skin on the arm of a Stretch robot.}
    \label{fig:teaser}
\end{figure}

We are inspired both by cane and walker design as well as how caregivers assist individuals with mobility challenges. We observe that information is conveyed through physical contact, in that users often apply specific grasping force patterns to a caregiver's arm to convey their intentions: a slight twist to indicate a turn or a pull to signal a stop  (Fig.~\ref{fig:teaser} left). To replicate this natural and intuitive dynamic, we propose a novel interface that uses a tactile sensor patch that functions like robotic skin, enabling the robot to ``feel" the user's nuanced touch gestures. This sensor is further integrated onto a cylindrical handle mounted on a Stretch 3 mobile manipulator (Fig.~\ref{fig:teaser} right). While a natural design for grasping the robot, determining human intent for this form factor proves challenging for traditional algorithms. To effectively classify the tactile data for human mobility intent recognition, we developed a novel \textit{Cylindrical Kernel Support Vector Machine (CK-SVM)}. This algorithm is designed to leverage the sensor's cylindrical geometry by enhancing the standard Radial Basis Function Support Vector Machine (RBF-SVM) to be robust against slight rotational shifts in the user's grasp, thereby increasing classification accuracy.
Finally, we integrated the sensor and CK-SVM into a complete RAN system named \textit{Tac-Nav}. We performed a comprehensive evaluation, benchmarking the CK-SVM against standard classifiers such as RBF-SVM and Convolutional Neural Network (CNN) using both simulated and real-world tactile data. Furthermore, we conducted a pilot study with 5 subjects to compare the performance and user experience of Tac-Nav against conventional joystick-based and voice-controlled RAN systems. The results demonstrate the superior classification accuracy of CK-SVM and highlight the benefits of Tac-Nav in providing a more effective and user-preferred navigation experience.

We present the following contributions:
\begin{itemize}
    \item A novel Cylindrical Kernel SVM (CK-SVM) designed for our tactile sensor, which improves its robustness to rotational shifts in the user's grasp.
    \item The Tac-Nav system, which integrates our sensor and algorithms on a mobile manipulator for intuitive, tactile-based assistive navigation.
    \item A comprehensive evaluation of CK-SVM against other classifiers using real and simulated data, and a pilot study comparing Tac-Nav to two other RAN systems.
\end{itemize}

\section{RELATED WORK}

\subsection{Robot Assistive Navigation}

Robot assistive navigation (RAN) systems can be broadly categorized based on their target users: those designed for individuals with visual impairments and those for individuals with mobility impairments.
For the visually impaired, the primary goal of a RAN system is to facilitate wayfinding to a target destination while ensuring obstacle avoidance \cite{pohovey2025beyondcanes}. For example, Balatti et al. employed a Franka manipulator on a mobile base with an impedance controller to achieve adaptive pulling \cite{Balatti2024RobotAN}. The DRAGON system mounted a handle on a TurtleBot and implemented a dialogue system to enable navigation through natural language commands \cite{DRAGON}. Other research has focused on augmenting the traditional white cane with advanced sensing and feedback capabilities \cite{Kebede2023AssistiveSC, Mai2023LaserSV, technologies12060075, Gad2023SafeNT}.
For individuals with mobility impairments, RAN systems are primarily aimed at providing physical support and reducing the effort required to perform tasks such as maintaining balance \cite{annurev:/content/journals/10.1146/annurev-control-062823-024352}. The ``NavChair" was one of the first shared-control wheelchair systems \cite{NavChair}, and its development was followed by numerous other smart wheelchair projects \cite{Simpson2005Smart, Leaman2017ACR}. Other prominent forms of RAN include robotic walkers \cite{Page_Saint-Bauzel_Rumeau_Pasqui_2017} and exoskeletons \cite{Siviy2023Opportunities}.
Research in RAN often explores various human input modes (e.g., joystick \cite{KIVRAK2021SocialNF}, brain-machine interface (BMI) \cite{Bellary2020IndoorNA}), feedback modalities (e.g., audio \cite{DRAGON}, haptics \cite{JIMENEZ202018}), and sensor integrations (e.g., camera, LiDAR, IMU \cite{Lu2021Assistive}). Notably, tactile sensing\footnote{It is worthwhile distinguishing between \textit{tactile input / tactile sensor}, through which the user conveys intent to the system via physical contact, and \textit{vibrotactile output / tactile actuator}, by which the system delivers feedback or alerts to the user via vibration.} has been underexplored. While prior work such as \cite{10.3389/fnbot.2020.575889} has used simple pressure sensors for singular tasks like fall detection, they lack the data richness required for complex command recognition. Similarly, force-torque sensors \cite{Ranganeni2023ExploringLC, Sutera2025AdaptiveNS} often require rigid, embodiment-specific mechanical integration that limits their versatility across different robots. In contrast, our soft tactile skin is universal and can be applied to diverse surfaces such as robot arms, custom handles, or even humanoid frames without extensive hardware modification, leaving its full potential for assistive navigation yet to be explored.

\subsection{Assistive Human Intent Recognition}
Accurate and efficient recognition of human intent is crucial for any assistive robot. Various algorithms have been proposed for different sensor types, yet limitations persist. The DRAGON system, for instance, maps voice commands to landmarks using a fine-tuned CLIP model, but its speech-to-text performance can degrade in noisy environments \cite{DRAGON}. Jain et al. proposed a recursive Bayesian filtering approach using non-verbal observations, but such probabilistic modeling can impose restrictive assumptions on features and environments~\cite{Jain2019ProbabilisticHI}. Shen et al. placed three resistive force sensors on a lower extremity assistive device and used Principal Component Analysis (PCA) followed by an SVM to classify motion intent, though this required specialized and expensive hardware \cite{Shen2013MotionIR}. Ranganeni et al. used a one-dimensional torque sensor to detect a user's turning intent, but the sensor could not provide richer contextual information \cite{Ranganeni2023ExploringLC}.
Alternatively, EEG sensors can provide rich brain signals. For example, Choi et al. used sample covariance matrices from EEG signals to classify grasping intent, and Perrin et al. used EEG to confirm or reject proposed actions~\cite{PERRIN20101246,Peng}. However, EEG signals are notoriously noisy, particularly when the user is in motion. Other work has explored deep learning models with multi-modal sensor inputs to learn underlying human intent \cite{SUN2025107254}. While these methods can achieve superior performance, their lack of interpretability can be a significant concern in safety-critical assistive applications. In this work, we achieve robust human intent recognition using a classical SVM enhanced with a novel, interpretable kernel, applied to high-dimensional tactile data.

\subsection{Tactile Gesture Classification}

To enable robots to understand human intent via touch, researchers typically employ gesture-classification methods, most often using a pressure-sensing grid as the sensing apparatus. Traditional approaches have frequently relied on classical machine learning models like boosting, random forests, and SVMs \cite{Gaus2015-ot, Flagg2013-gf}. For instance, the CoST social gesture dataset was classified using SVMs and Bayesian classifiers on extracted features such as mean pressure and gesture duration \cite{Jung2014-am}. Similarly, other work has used random forests to classify different types of infra hug gestures \cite{Block2023-qm}.
While these methods are still in use, there is a growing trend toward using deep learning to classify touch gestures \cite{jung_2024, Yang2022-yj, Albawi2018-wz, Crowder2025-ge}. Some of these advanced methods enhance recognition by extracting spatiotemporal features from wavelet coefficients \cite{li_2022}, and another novel approach used microphone arrays as a form of robotic skin, feeding spectrograms into a convolutional neural network \cite{Yang2024-is}. Although numerous methods have been developed, they are often highly specialized for the specific gestures and sensor modalities for which they were designed.

\begin{figure*}[!t]
    \centering
    \includegraphics[width=1\textwidth]{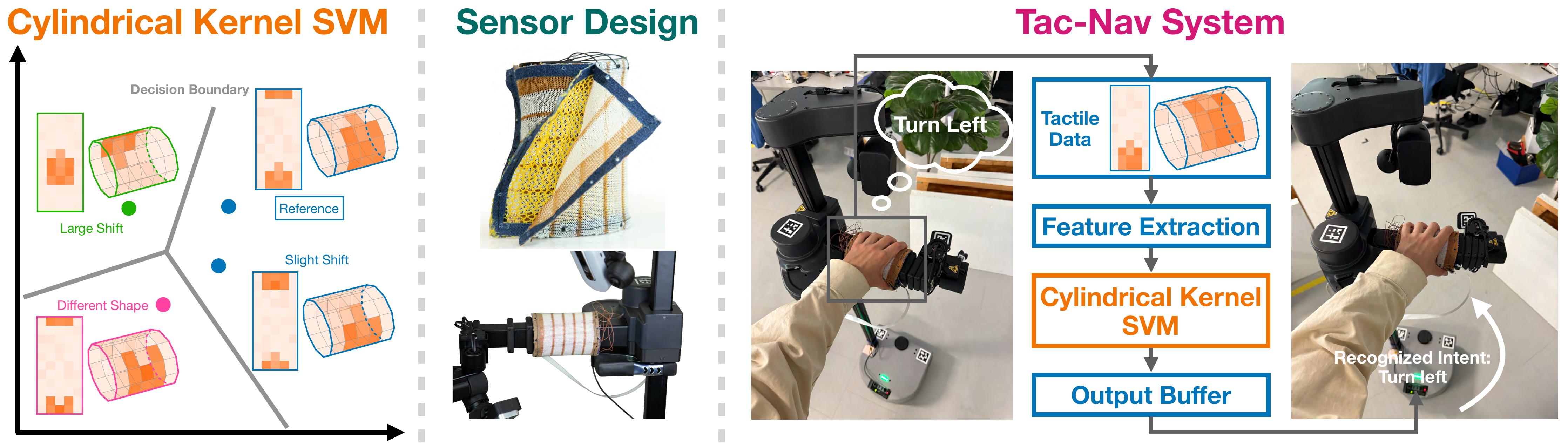} 
    \caption{\textbf{Left: CK-SVM illustration.} The CK-SVM classifies tactile patterns based on their similarities. Given a \textcolor{_blue}{\textbf{reference}} tactile data, \textcolor{_blue}{\textbf{patterns with slight shifts}} are classified to be the same class, while either \textcolor{_green}{\textbf{patterns with large shifts}} or \textcolor{_pink}{\textbf{data with different patterns}} are classified to be different classes. Draw in 2D for visualization simplicity. \textbf{Middle: Sensor Deign.} The three-layer knitted tactile sensor and its mounting on the arm of the Stretch 3 robot. \textbf{Right: Tac-Nav system overview.} For a given intent such as ``turn left", the system reads tactile data, extracts features that preserve the cylindrical topology, classifies the features using the CK-SVM, and sends the final recognized intent to the robot after filtering by an output buffer.}
    \label{fig:main} 
\end{figure*}

\section{METHOD and SYSTEM}

This section details our methodology, beginning with CK-SVM, our main algorithmic contribution. We then describe the implementation of the complete Tac-Nav system, covering the sensor design and hardware configuration in Section~\ref{sec: sensor design}. This is followed by a description of the software pipeline — including data collection, model training, and real-time control logic — in Section~\ref{sec: software}.

\subsection{Cylindrical Kernel SVM}
\subsubsection{Preliminary -- Support vector machine with radial basis function kernel (RBF-SVM)}
\label{sec: RBF-SVM}
The Support Vector Machine (SVM) is a powerful supervised learning algorithm, and its effectiveness can be extended to non-linearly separable datasets through the use of the kernel trick \cite{SVM}. A popular and effective choice for this is the Radial Basis Function (RBF) kernel, which implicitly maps input features into a high-dimensional space where a linear separation becomes feasible. This kernel function measures the similarity between two feature points $x_1$ and $x_2$ using a Gaussian function, defined as:
\begin{equation}
K(x_1, x_2) = \exp(-\gamma \cdot ||x_1 - x_2||^2)
\end{equation}
In this equation, $||x_1 - x_2||^2$ represents the squared Euclidean distance between the two feature vectors. The hyperparameter $\gamma$ acts as the length-scale, controlling the ``spread" or influence of the kernel.

With the kernel defined, the SVM decision function for classifying a new sample feature $x$ is formulated as:
\begin{equation}
y = \text{sign}\left(\sum_{i=1}^{N} (\alpha_i \cdot y_i \cdot K(x, x_i)) + b\right)
\end{equation}
where $y$ is the predicted label, $N$ is the number of support vectors, $\alpha_i$ are the learned Lagrange multipliers, $y_i$ are the corresponding class labels of the support vectors, and $b$ is the bias term. This function computes a weighted sum of the kernel-based similarities between the input sample and the support vectors to make a final classification. In the context of this work, it maps complex tactile features to their corresponding labels: ``turn left", ``turn right", ``speed up", ``stop", and ``neutral".

\subsubsection{Cylindrical Kernel SVM (CK-SVM)}
Different human intents correspond to distinct grasping position and local patterns across the tactile handle. However, because we do not restrict human holding gesture and there is grasping position inconsistency across users and trials, these patterns are subject to rotational shifts along the cylindrical perimeter of the handle, even when the intended action is the same. 
If a standard SVM classifier with an RBF kernal is used on user data from the cylindrical form factor of our tactile sensor (as shown in Fig.~\ref{fig:main} and explained in Section~\ref{sec: sensor design}), intent recognition is not reliable (accuracy of approximately 0.37 for five intent classes). An analysis of this result revealed the following key observation: 

\begin{quote}
\textit{The standard kernel fails because its ``flattening" operation destroys the tactile sensor's cylindrical topology, which causes it to misinterpret the slight rotational shifts in a user's grasp as large, unrelated data changes.}
\end{quote}

To be interpreted correctly, this tactile feature should be viewed on its native 3D cylindrical manifold. A standard RBF kernel, however, requires the feature to be ``flattened" into a one-dimensional vector. This operation destroys the essential 3D topology, causing a minor rotational shift -- particularly the one that ``\textit{wraps around}" from the last row of the sensor to the first -- to make two otherwise similar grasp patterns appear drastically different in their vector representations. Although the change is insignificant on the sensor's 3D surface, the large Euclidean distance created in the flattened space leads the RBF kernel to misinterpret the patterns as highly dissimilar, resulting in poor classification performance.

To address the misinterpretation of rotational shifts, we introduce the \textbf{Cylindrical Kernel}, a modification of the standard RBF kernel designed to respect the sensor's cylindrical topology. Our kernel retains the Gaussian form but replaces the Euclidean distance with a custom \textbf{Cylindrical Distance} metric, $d_C(x_1, x_2)$:

\begin{equation}
K(x_1, x_2) = \exp(-\gamma \cdot d_C(x_1, x_2))
\end{equation}

The Cylindrical Distance finds the minimum cost to align two tactile feature maps, $x_1$ and $x_2$, by rotationally shifting one relative to the other:

\begin{equation}
d_C(x_1, x_2)=\min_{s\in{0, ..., k-1}} (\|x_1-\tau_s(x_2)\|_F^2+\Lambda(s))
\end{equation}

This metric consists of two key components:
\begin{enumerate}
    \item $\|x_1-\tau_s(x_2)\|_F^2$: The squared Frobenius distance between the first feature map and the second map after it has been rotationally shifted by $s$ rows.
    \item $\Lambda(s)$: A penalty function that discourages large shifts.
\end{enumerate}

The shift operation and penalty function are defined as follows:
\begin{equation}
\label{eq: tau}
[\tau_s(x)]_{i,j} \equiv x'_{i,j} = x_{((i-s)\text{ mod } k),j}
\end{equation}
\begin{equation}
\label{eq: lambda}
\Lambda(s)=\exp(\min(s,k-s)/\delta)-1
\end{equation}

Equation \ref{eq: tau} details the circular shift for each element at row $i$ and column $j$ of the feature map, where $k$ is the total number of rows on the sensor. Equation \ref{eq: lambda} defines the exponential penalty for a given shift $s$, where the $\min(s, k-s)$ term measures the shortest rotational distance around the cylinder. This penalty ensures that two patterns are considered different even if they match perfectly after a large rotational shift, preventing misclassifications between similar grasp patterns that are far apart on the handle. $\delta$ is a tunable hyperparameter.

By explicitly capturing the cylindrical geometry of the tactile sensor, our proposed CK-SVM is theoretically poised to increase classification accuracy. We validate this hypothesis through a comprehensive evaluation of CK-SVM on both simulated and real-world data, presented in Section \ref{sec: eval}.

\subsection{Tac-Nav System Implementation}

\subsubsection{Hardware Configuration}
\label{sec: sensor design}
To obtain user tactile data, a machine knitted resistive-based sensor was created. Using machine knitting allows for quick and cheap sensor fabrication allowing for a faster sensor iteration cycle. Since people touch fabric every day, creating a sensor using a similar material makes the sensor more comfortable and familiar. 

Our sensor adopts the three-layer architecture proposed by Si et al. \cite{Si2023}, comprising two conductive layers patterned with orthogonal stripes and a central non-conductive mesh. The resulting sensing array spans a 11 × 5 grid, offers a spatial resolution of approximately 15.5 mm$^2$, and operates at a sampling rate of about 50 Hz. The sensor has a height of 15 cm and a diameter of 7.4 cm. The inner mesh was created with A and E nylon yarn and the insulating stripes between each conductive strip is made of rayon yarn. The conductive stripes are composed of stainless steel infused yarn which had a resistance of around 2.5M$\Omega$. A NUCLEO-F030R8 board was used.

Then, this tactile sensor array is mounted on a 3D-printed cylindrical holder. This assembly is affixed to the horizontal arm of a Hello Robot Stretch 3, as shown in Figure~\ref{fig:main}. Mounting the sensor on the adjustable arm, rather than the main vertical body, offers the ergonomic benefit of allowing users to easily set the handle to their preferred height. To ensure a clean and unobtrusive setup, the control board is housed underneath the arm, with wiring routed along the arm to the main body.

\subsubsection{System Pipeline}
\label{sec: software}
Our system is designed to recognize a set of five human intents: \{``turn left", ``turn right", ``speed up", ``stop", ``neutral"\}. The software pipeline, from data collection to real-time control, is as follows:

\paragraph{Data Collection and Training}
\label{sec: data collection}
To minimize user burden, we collected a concise dataset for each participant, consisting of 40 one-second samples for each of the five intents. The users are encouraged to use their natural holding pose for each intent instead of a prefixed grasping pose to account for distinct user preferences. By fixing the sampling frequency to be 45 Hz, the preliminary dataset for each subject has a shape of $(200,45,11,5)$. From each one-second sample, we extract features including the mean, max, standard deviation, and spatial gradient of the tactile pressures, which collapses the time-series data into four 2D feature maps that preserve the sensor's cylindrical topology, resulting in a final dataset for each subject of shape $(200,4,11,5)$. Then we shuffle the dataset and split 80\% to be the training set, and the rest for test set. Our CK-SVM model is then trained on the training set, with a total training time of less than four seconds per subject.

\paragraph{Real-Time Classification} During operation, the system uses a sliding window approach for real-time intent classification. A one-second window, consistent with the training data, slides in 100-millisecond increments. To enhance system robustness and prevent reactions to occasional misclassifications, we implement an \textit{output buffer} that stores the latest seven predictions. A new command is sent to the robot only when all seven intent classification results in the buffer are identical; otherwise, a ``neutral" command is issued, which either stops turning or maintains the current moving velocity.

\paragraph{Robot Control} We use ROS2 to communicate the final classified intent to the Stretch 3 robot. For robot motion, the turning angular speed is set to 0.15 radians per second, and each ``speed up" command increases the robot's forward velocity by 0.01 meter per second, up to a maximum of 0.15 meter per second. The ``stop" command brings the robot to an immediate halt.

\begin{figure*}[!t]
  \centering
  \includegraphics[width=1\textwidth]{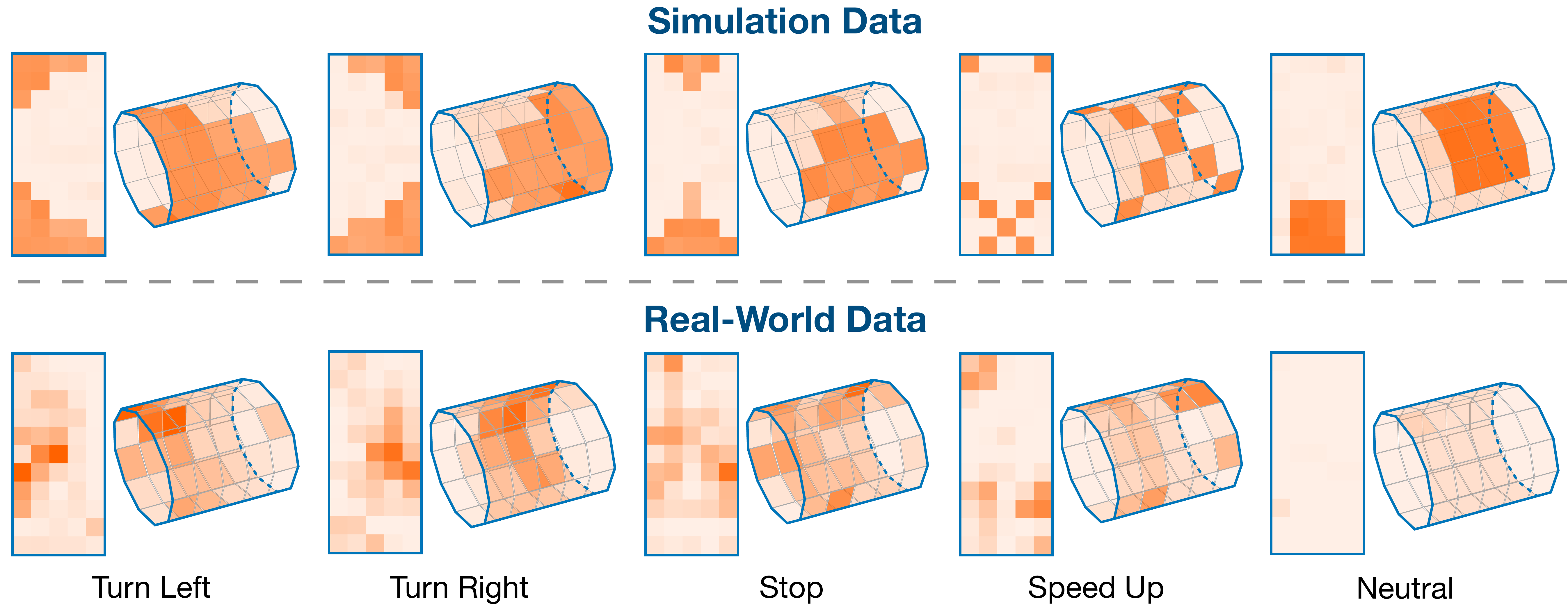}
  \caption{\textbf{Tactile data visualization for all intents.} Top: designed simulated base patterns. Bottom: mean patterns across trials from one user.}
  \label{fig:pattern_vis}
\end{figure*}

\begin{table*}[htbp]
\centering
\caption{\textnormal{\small{Classification accuracies $(\text{mean} \pm \text{std})$ of four baseline methods and CK-SVM on simulated and real-world data.}}}
\begin{tabular}{lccccc} 
\toprule
& RBF-SVM & MLP & MDCM & CNN & \textbf{CK-SVM} \\ \midrule
Simulated Dataset & $0.7708 \pm 0.0579$ & $0.7375 \pm 0.0818$ & $0.7417 \pm 0.0847$ & $0.8125 \pm 0.0209$ & $\mathbf{0.9708} \bm{\pm} \mathbf{0.0188}$ \\
Real-World Dataset & $0.3708 \pm 0.0219$ & $0.3807 \pm 0.0113$ & $0.5683 \pm 0.1250 $ & $0.7599 \pm 0.0782$ & $\mathbf{0.9081} \boldsymbol{\pm} \mathbf{0.0198}$ \\ \bottomrule
\end{tabular}
\label{tab}
\end{table*}


\section{INTENT CLASSIFICATION EVALUATION}
\label{sec: eval}

\begin{figure}[!htbp]
    \centering
    \includegraphics[width=0.8\columnwidth]{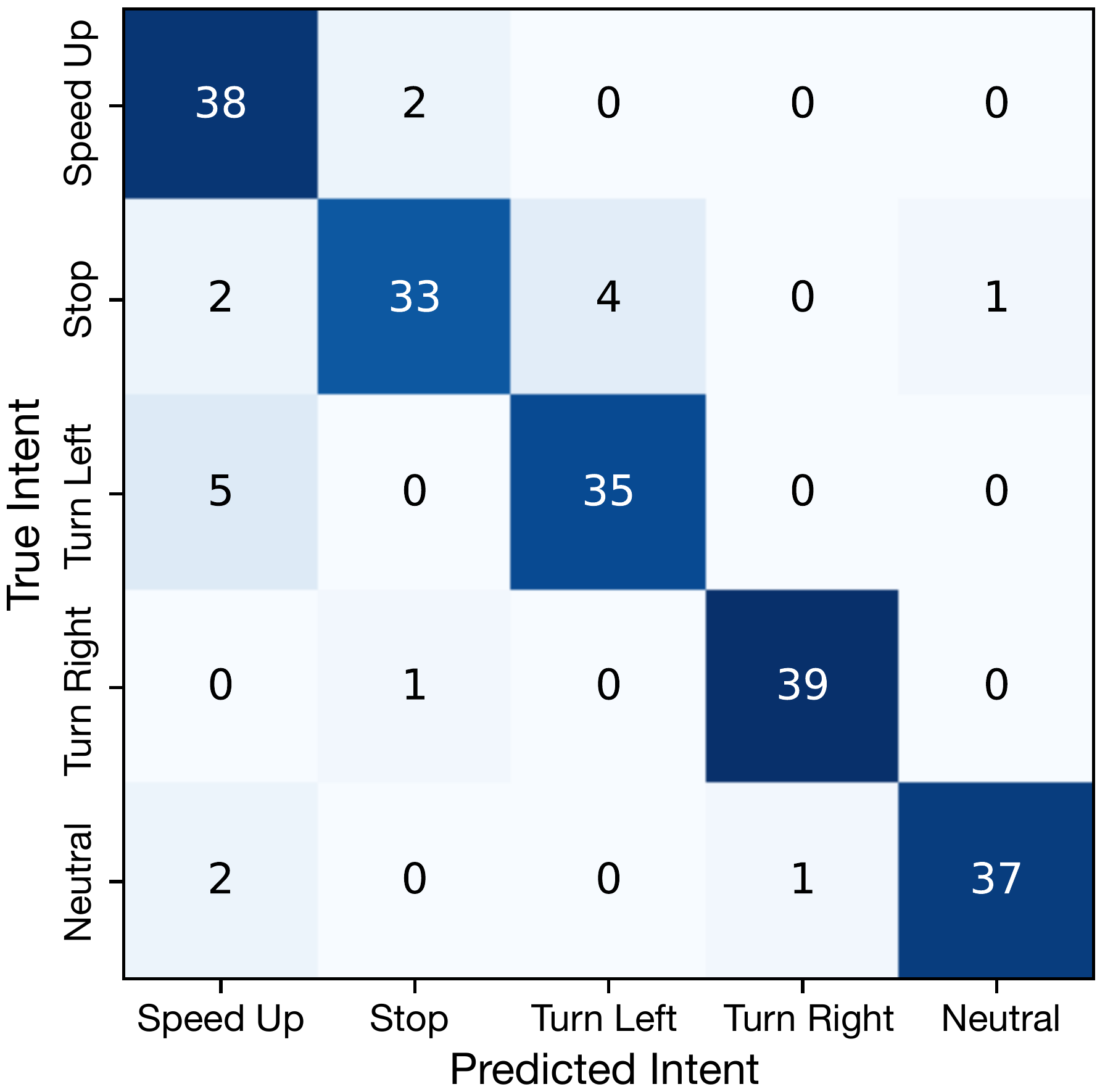} 
    \caption{Aggregated confusion matrix for CK-SVM on real-world dataset, summing the individual ones from all subjects.}
    \label{fig:confusion} 
\end{figure}

To evaluate the performance of our proposed CK-SVM, we compare it against the following four baseline classifiers: 
\begin{enumerate}
    \item \textbf{Standard SVM with an RBF kernel.} The RBF-SVM introduced in Section~\ref{sec: RBF-SVM} is chosen to provide a direct comparison with our Cylindrical Kernel, isolating the benefit of our geometry-aware design.  
    \item \textbf{Multilayer Perceptron (MLP).}  The MLP is selected as a powerful, general-purpose non-linear classifier. 
    \item \textbf{Minimum Distance to Covariance Mean (MDCM).} MDCM is a common method for biosignal classification, making it a good candidate because tactile sensor data can be viewed as a multi-channel time-series signal. Specifically, for each class $k$, a sample covariance matrix centroid $M^k$ is calculated from the training data. For a new data sample $X$, the predicted class $\hat k$ is given by the nearest centroid: $\hat k = \arg\min_k d(X,M^k)$, where the distance metric $d$ is the geodesic distance between two sample covariance matrices on their Riemannian manifold. Please refer to \cite{Barachant2012MulticlassBCI} for more details. 
    \item \textbf{Convolutional Neural Network (CNN).} 
    The CNN is chosen as a strong baseline due to its inherent ability to learn local spatial patterns directly from the 2D tactile data. However, it lacks interpretability and typically requires a larger dataset to train effectively and avoid overfitting, which may increase human burden.
\end{enumerate}

We evaluate the overall classification accuracies of all methods on both simulated dataset and real-world dataset. Based on the later dataset we also provide the confusion matrix to gain deeper insights into our models.

\begin{figure*}[!t]
    \centering
    \includegraphics[width=1\textwidth]{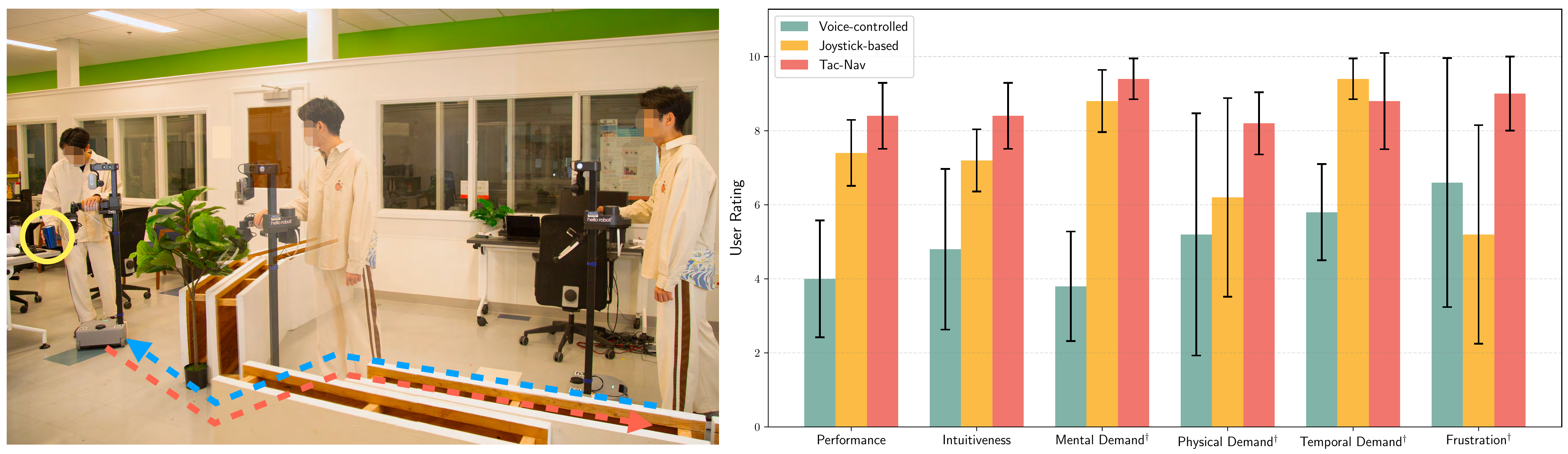} 
    \caption{\textbf{Left: Experiment setup.} The user starts from right, follows the blue arrow to pick up the water bottle on the left, and returns by following the red arrow. \textbf{Right: Pilot study results.} To unify the visual representation of our metrics, ``mental demand", ``physical demand", ``temporal demand", and ``frustration" were inverted, where a higher score indicates a more positive outcome (e.g., easy to use).}
    \vspace{-10pt}
    \label{fig:exp} 
\end{figure*}

\subsection{Simulated Data Results}

To validate our method in a controlled setting, we first conducted a proof-of-concept experiment on simulated data. We generated a synthetic dataset by creating five distinct local patterns, each corresponding to one of the five intents, shown as the top row of Fig.~\ref{fig:pattern_vis}. The final dataset was produced by applying random rotational shifts and adding Gaussian noise. The performance of the baseline methods and our CK-SVM on this dataset is presented in the first row of Table~\ref{tab}. While all classifiers achieved respectable accuracies above 0.73, our CK-SVM demonstrated superior performance, attaining a classification accuracy of 0.9708 with the lowest standard deviation of 0.0188. This result highlights its robustness to the specific geometric challenges it was designed to overcome.

\subsection{Real-World Data Results}

We next evaluated the classifiers on real-world data collected from five human subjects, following the protocol described in Section~\ref{sec: data collection}. To provide an intuitive understanding of this data, the bottom row in Fig.~\ref{fig:pattern_vis} visualizes the mean tactile activation maps for each intent, averaged across all trials from one user. The visualizations reveal intuitive patterns; for instance, the left side of the sensor shows higher activation for the ``turn left'' intent, while the right side is more active for ``turn right''. However, as expected, these real-world patterns are substantially more complex and less distinct than those in the simulated dataset.
The quantitative results, summarized in Table~\ref{tab}, show a general performance decrease compared to the simulated data due to this increased complexity. The RBF-SVM, MLP, and MDCM models struggled, achieving accuracies below $0.6$. While the CNN performed better with an accuracy over $0.75$, its large standard deviation indicates inconsistent performance across subjects. In contrast, our CK-SVM achieved the highest classification accuracy of over $0.9$ with a low standard deviation of $0.02$, demonstrating both the superior performance of our cylindrical kernel and its robustness across different users. To gain some deeper insights of the CK-SVM's performance, we summed over the classification results and the corresponding true labels for all users, and provided the aggregated confusion matrix as shown in Fig.~\ref{fig:confusion}.
This matrix reveals that the ``speed up'' intent has a noticeably low precision of approximately 0.81. We attribute this to the physical nature of the interaction, as the user's palm remains in constant contact with the sensor; this means that even during a turning or ``stop'' action, some forward pressure is often present, leading to a higher rate of false positives for the ``speed up'' class.

\section{REAL-WORLD NAVIGATION EXPERIMENTS}

Beyond offline classification accuracy, we evaluated the complete Tac-Nav system by conducting a pilot study with five subjects. This approach allowed us to gain a deeper understanding of the system's real-world performance and usability by collecting a range of user experience metrics.

\subsubsection{Experimental Setup}
The experiment for each user was conducted in two phases, illustrated as the left image of Fig.~\ref{fig:exp}. Before the experiments, participants were asked to imagine they had limited mobility, so they would rely on the Stretch 3 robot as the form of a cane or a walker, similar to an older adult requiring such walking balance assistance. After the initial data collection and model training period as introduced in Section ~\ref{sec: data collection}, users are required to complete a navigation task mimicking a common in-home scenario: navigating to a location to retrieve a cup of water and returning. In our setup, the users start from the right side of the image, follow the right-to-left blue arrow to reach the table on the left, get the blue water bottle circled in yellow, and follow the left-to-right red arrow to get back to the starting place. The path was designed to include a narrow passage that necessitated both left, right turns and stop. For the return journey, users would have to hold the water bottle while walking and maintain balance.

\subsubsection{Comparisons \& Evaluation Metrics}
To evaluate the performance of our Tac-Nav system, we designed two alternative control systems for comparison: one based on natural language voice commands and another using a joystick.

For the voice-controlled system, we use python \textit{SpeechRecognition} as voice-to-text model. A ``turn left'' or ``turn right'' instruction causes the robot to stop and turn 15 degrees in the specified direction. The ``speed up'' command increases the robot's velocity by 0.07 meter per second, with a maximum speed of 0.14 meter per second. The ``stop'' command brings it to a complete halt. For the joystick-based system, the controller was mounted on the robot's shoulder for easy access with the user's left hand, allowing their right hand to hold the robot's arm for balance. The control logic was standard: pushing forward commanded forward motion at a velocity of 0.1 meter per second, while left and right inputs produced an angular velocity of around 30 degree per second. Pushing backward was disabled, and releasing the joystick stopped the robot.

User experience with all three systems was evaluated using a questionnaire based on the NASA Task Load Index (TLX) \cite{Hart1988NASA}, which assesses ``mental demand'': How mentally demanding the task was, ``physical demand'': how physically demanding the task was, ``temporal demand'': how hurried or rushed the pace of the task was, ``performance'': how successful the user was in accomplishing what
she was asked to do, and ``frustration'': how insecure, discouraged, irritated, stressed, and annoyed the user was. We augmented this with an ``intuitiveness'' metric to measure how natural and intuitive the user perceived each interface for controlling the robot. After using all three systems, participants provided 10-point scale ratings for each metric, followed by a short interview to gather qualitative feedback.

\subsubsection{Pilot Study Results}
The pilot study results are presented in Fig.~\ref{fig:exp}. The voice-controlled system, joystick-based system, and our Tac-Nav system are colored in green, yellow, and red, respectively. The error bars represent the standard variance across all participants. The `$\dagger$' sign after ``mental demand", ``physical demand", ``frustration", and ``temporal demand" indicates the scores are inverted for a unified metric visual representation, where a higher score indicates a higher user preference (i.e., low demand or frustration).

The results show a strong user preference for our Tac-Nav system. It outperformed both the joystick-based and voice-controlled systems in five out of the six metrics, exhibiting the highest average scores and the lowest variance across subjects, which demonstrates its superior overall performance and consistent user experience.
Notably, for ``physical demand'' and ``frustration'', Tac-Nav scored more than two points better than the alternatives. Also, the large standard variances of joystick-based system reflect a split in user experience: while two participants who successfully complete the task found the task to be manageable, three users commented that using the joystick while holding the cup of water was physically demanding and frustrating, especially when water was occasionally spilled. In contrast, our Tac-Nav system provided a more natural interaction, enabling users to simultaneously control the robot's moving direction and maintain balance with one hand while carrying the water bottle with the other.
The only metric where the joystick performed slightly better was ``temporal demand'' (by 0.5 points). This is attributable to the joystick's direct real-time control, whereas the output buffer in Tac-Nav introduces a minimum delay of 0.2 seconds, with occasional classification noise at the start of a movement increasing the stabilization time to nearly one second. However, all five subjects stated this delay was acceptable, especially given the significant improvement in reliability that the buffer provided. Regarding the voice-based controller, four out of five users found the speech recognition to be imperfect and noted that issuing frequent verbal commands, particularly for large turns, could be tedious.

\section{CONCLUSION and DISCUSSION}

\subsection{Conclusion}
Inspired by the way caregivers interpret physical grasping cues from those they assist, this work presents Tac-Nav, a robot assistive navigation system that uses a tactile skin interface for intuitive human intent recognition. To address the challenge of natural variations in a user's grasp, we developed the Cylindrical Kernel Support Vector Machine (CK-SVM), an algorithm specifically designed to be robust to rotational shifts by explicitly modeling the sensor's cylindrical topology. Integrated into a real-time system with a filtering output buffer, the CK-SVM enables Tac-Nav to provide fluid and robust mobility support based on the user's touch.
Intent classification evaluation results demonstrate the superiority of our approach, with the CK-SVM achieving highest classification accuracy than baseline models on both simulated and real-world data. Furthermore, a real-world navigational pilot study confirmed the high usability and user preference for Tac-Nav over conventional voice-controlled and joystick-based systems, validating the effectiveness of our tactile-based system.

\subsection{Limitations and Future Work}
While this work demonstrates the benefits of our tactile-based robot assistive navigation system, we acknowledge several limitations that inform our future work. First, our current CK-SVM is specifically designed for cylinder-shaped sensors, which we will explore generalizing our kernel design to other common handle geometries. Besides, our findings on user preference are preliminary based on a small pilot study. In the future we will expand this into a comprehensive clinical trial with mobility-impaired individuals to validate the system's long-term usability. Finally, the rich data from the tactile skin is currently only used for mobility intent recognition of five classes; we intend to leverage its full potential to enable new functionalities, such as inferring the user's physical state (e.g., fatigue or confidence) from their grip forces and patterns, or enhancing safety by detecting unstable grips indicative of a potential fall.

\section*{ACKNOWLEDGMENT}
This work was supported by the National Science Foundation under Grant CCF 2236484. We would like to thank Haonan Chen for his valuable feedback during proofreading.




\bibliographystyle{IEEEtran}
\bibliography{references.bib}

\end{document}